\documentclass[journal,10pt]{IEEEtran}
\usepackage{url}

\usepackage{caption}
\usepackage{subfig}
\usepackage{times}
\usepackage{epsfig}
\usepackage{graphicx}
\usepackage{amsmath}
\usepackage{amssymb}
\usepackage{float}
\usepackage{multirow}
\usepackage{algorithm}

\usepackage{algorithmic}

\ifCLASSINFOpdf
\else
\fi
\usepackage{amsmath}
\begin{document}
%
\title{Adaptive Period Embedding for Representing Oriented Objects in Aerial Images}
%
%

\author{Yixing Zhu, Xueqing Wu, Jun Du\\
  National Engineering Laboratory for Speech and Language Information Processing\\ University of Science and Technology of China\\
  Hefei, Anhui, China\\
  {\tt\small zyxsa@mail.ustc.edu.cn, shirley0@mail.ustc.edu.cn, jundu@ustc.edu.cn}
}

\maketitle

\begin{abstract}
We propose a novel method for representing oriented objects in aerial images named Adaptive Period Embedding (APE). While traditional object detection methods represent object with horizontal bounding boxes, the objects in aerial images are oritented. Calculating the angle of object is an yet challenging task. While almost all previous object detectors for aerial images directly regress the angle of objects, they use complex rules to calculate the angle, and their performance is limited by the rule design. In contrast, our method is based on the angular periodicity of oriented objects. The angle is represented by two two-dimensional periodic vectors whose periods are different, the vector is continuous as shape changes. The label generation rule is more simple and reasonable compared with previous methods. The proposed method is general and can be applied to other oriented detector. Besides, we propose a novel IoU calculation method for long objects named length independent IoU (LIIoU). We intercept part of the long side of the target box to get the maximum IoU between the proposed box and the intercepted target box. Thereby, some long boxes will have corresponding positive samples. Our method reaches the $1^{\text{st}}$ place of DOAI2019 competition task1 (oriented object) held in workshop on Detecting Objects in Aerial Images in conjunction with IEEE CVPR 2019.

\end{abstract}


%
\IEEEpeerreviewmaketitle

\section{Introduction}
\begin{figure}
  \begin{center}
       \includegraphics[width=1.0\linewidth]{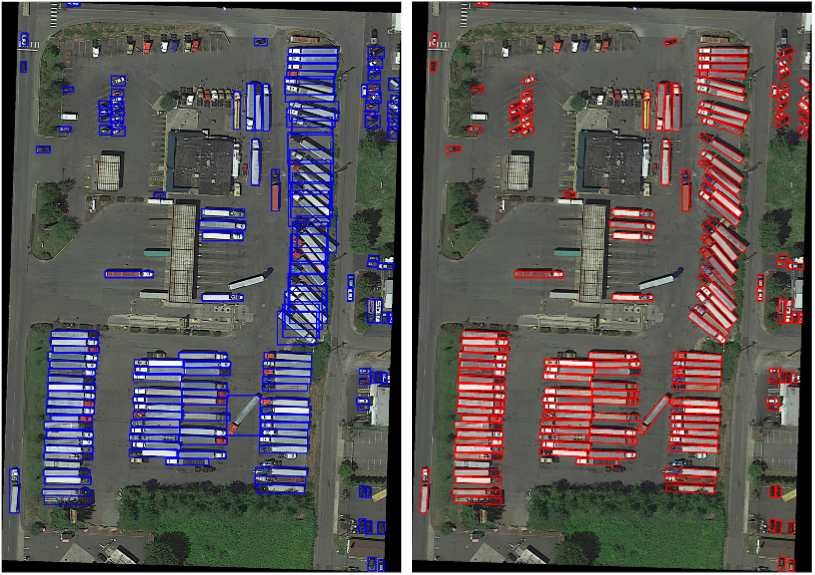}
    \end{center}
    \caption{Left: horizonal bounding boxes, right: oriented bounding boxes.
  }
  \label{hori_ori}
\end{figure}

Traditional object detections mainly detect objects with horizontal bounding boxes. However, objects in aerial images are oriented and cannot be effectively represented by horizontal bounding boxes. As shown in Fig. \ref{hori_ori}, detecting oriented objects with horizontal bounding boxes will contain more background and cannot accurately locate the objects. Besides, overlap calculation based on horizontal bounding box is not accurate for oriented objects, as the overlap between horizontal bounding boxes of two oriented objects may be too large, so NMS based on horizontal bounding boxes is not suitable for oriented objects. Thus, representing oriented objects with oriented bounding box is necessary for object detection in aerial images. Regressing oriented bounding box is more challenging than regressing horizontal bounding box. Four variables can represent a horizontal bounding box, such as x,y coordinates of top left corner and bottom right corner. However, oriented bounding box representation needs an extra variable $\theta$ to represent its angle. It is hard to directly regress $\theta$ because the angle is periodic.

Most of previous oriented detectors \cite{he2017deep} \cite{liao2018textboxes++} \cite{ma2018arbitrary} \cite{zhou2017east} \cite{ding2018learning} directly regress $\theta$ or the four vertices of the oriented bounding box and the label is calculated by complicated rules, which is hard for the network to learn. Some methods try to design a simple label calculation rule for oriented objects. For example, \cite{dai2017fused} adopts Mask R-CNN \cite{he2017mask} for detecting oriented text lines. \cite{zhu2018sliding} regresses the outline of objects with multiple points on sliding lines. But these methods introduce additional parameters and cannot be adopted by RPN (region proposal networks).

In this study, we propose a novel method for representing oriented objects. Oriented bounding box can be represented by ($x$, $y$, $w$, $h$, $\theta$), where $x$, $y$ are the coordinates of the center of the bounding box, and $w$, $h$ are the lengths of the long and short sides, respectively. We do not directly regress $\theta$. Instead, $\theta$ is represented by two two-dimensional periodic vector. The proposed method is different from \cite{long2018textsnake}, as in our method, the two vectors' periods are $90^{\circ}$ and $180^{\circ}$ respectively. Finally, we calculate the angle with these vectors. Our method is versatile and can be applied in other detectors. Besides, we design a novel cascade R-CNN method for long objects such as harbors. Generally, a two-stage model firstly proposes bounding boxes with RPN, and the output bounding boxes of the second step (R-CNN) are limited by RPN's results. Due to the limited size of the receptive field, some long objects cannot be covered by RPN. With this in mind, we adopt a two-stage cascade R-CNN model with length independent IoU (LIIoU) to detect long objects. In the first stage, some bounding boxes which only cover part of the objects are also set to positive samples. In this way, the first R-CNN can propose longer bounding boxes. The main contributions of our work are summarized as follows:

1. We present a novel method for representing oriented bounding boxes in aerial images. We do not directly regress $\theta$ of oriented bounding boxes, but instead embed $\theta$ with vectors whose periods are different. In this way, we do not need complex rules to label the angle which avoids ambiguity.

2. We present a novel IoU calculation method named length independent IoU (LIIoU), which is designed for long objects. The presented method makes the detector more robust to long objects.

3. The presented method achieves state-of-the-art on DOTA and wins the first place of Challenge-2019 on Object Detection in Aerial Images task1 (oriented task).

\section{Related Work}

\subsection{Horizontal objects detection}

Labels of traditional object detection tasks are horizontal bounding boxes. \cite{ren2017faster} presents a real-time object detection method based on region proposal networks (RPN) which shares feature maps of RPN and R-CNN and use anchors with different sizes and aspect ratios in RPN stage. Though Faster R-CNN shares feature maps, it still requires much computation in R-CNN's fully connected layer. Region-based fully convolutional networks (R-FCN) \cite{dai2016r} presents Position-sensitive score maps and Position-sensitive RoI pooling for saving computation in R-CNN stage. Scale variation is always
a very challenging issue in object detection; to help solve this problem, \cite{lin2017feature} presents Feature Pyramid Network (FPN). FPN generates feature maps of different scales on different layers, and detects large objects on higher layers but detects small objects on lower layers; the parameters of RPN is shared over layers. Based on FPN, Mask R-CNN \cite{he2017mask} presents RoIAlign which calculates values in RoI via bilinear interpolation instead of maximum to avoid quantization errors, and add several convolution layers on mask-head to generate instance segmentation maps. \cite{liu2018path} improves Mask R-CNN by adding Bottom-up Path Augmentation and feature fusion.

Two-stage methods require more computation than one-stage methods, so one-stage methods are more suitable for real-time object detection tasks. Single shot multibox detector (SSD) \cite{liu2016ssd} generates multiple layers, and then detect objects with different sizes on different layers. Deconvolutional single shot detector (DSSD) \cite{fu2017dssd} upsamples feature maps and detects small objects on lower layers which improves SSD's performance for small objects. \cite{lin2017focal} presents focal loss to handle the imbalance between positive and negative samples. Although anchors are widely used in object detection, many models adopt anchor-free method. \cite{huang2015densebox} does not use anchors in RPN, but uses a shrunk segmentation map as the label. \cite{redmon2016you} also uses segmentation maps as ground truth. GA-RPN \cite{wang2019region} combines anchor-free and anchor-based ideas: the label for the first step is generated by a shrunk segmentation map, and the label for the second step is calculated based on the output anchor of the first step.

\begin{figure*}
  \begin{center}
       \includegraphics[width=1.0\linewidth]{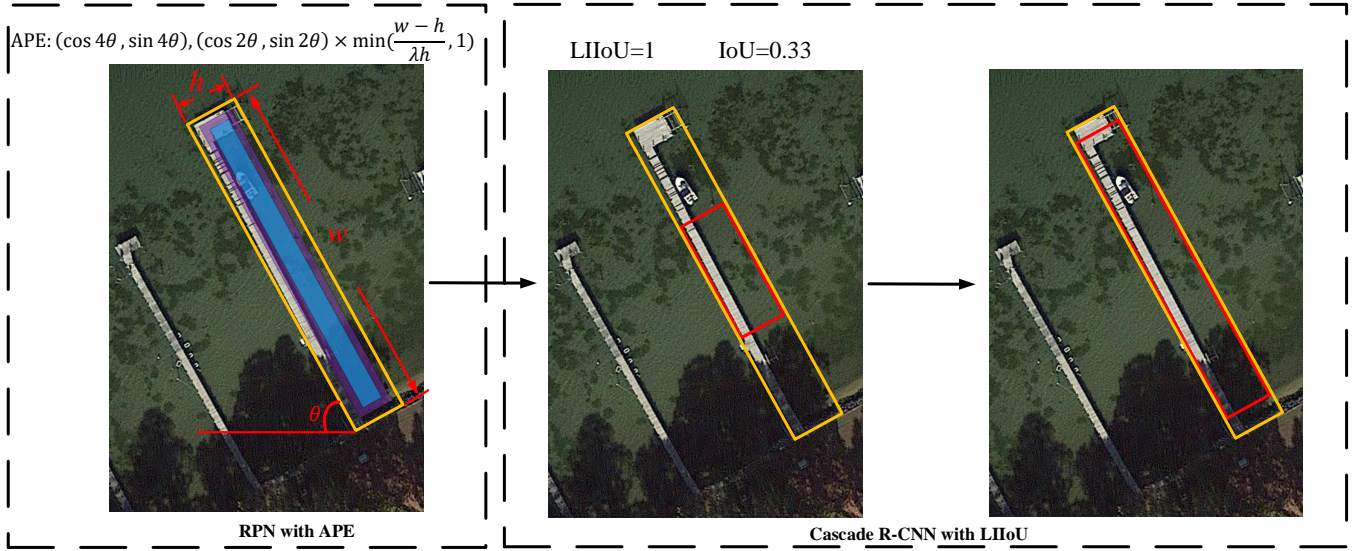}
    \end{center}
    \caption{Illustration of our proposed architecture. From left to right: Anchor-free RPN, Cascade R-CNN. Yellow bounding box is groud truth, red bounding box is proposed box. 
  }
  \label{framework}
\end{figure*}

Traditional object detection in aerial images only focuses on horizontal bounding box. \cite{li2018rotation} presents local-contextual feature fusion network which is designed for remote sensing images. The RPN includes multiangle, multiscale and multiaspect-ratio anchors which can deal with the oriented objects, but the final output bounding box is still horizontal. \cite{wang2017feature} presents rotation-invariant matrix (RIM) which can get both the angular spatial information and the radial spatial information. \cite{long2017accurate} presents an automatic and accurate localization method for detecting objects in high resolution remote sensing images based on Faster R-CNN. \cite{Arnt2015Detection} presents a method to detect seals in aerial remote sensing images based on convolutional network. \cite{chen2014vehicle} presents a hybrid DNN (HDNN), where the last convolutional and max-pooling layer of DNN is divided into multiple blocks, so HDNN can generate multi-scale features which improves the detector's performance for small objects. Unlike the images used for general object detection, the aerial image has a large resolution. However, large models cannot be implemented due to limited memory. So \cite{accurate2019object} proposes a self-reinforced network named remote sensing region-based convolutional neural network (R2-CNN) including Tiny-Net and intermediate global attention blocks. It adopts a lightweight residual structure, so the network can feedward huge resolution sensing images at high speeds. \cite{deng2019learning} proposes a novel method for ship detection in synthetic aperture radar (SAR) images. It redesign the network structure, does not pre-train on ImageNet, and specifically design the system for small objects such as ships.

\subsection{Oriented objects detection}

Oriented object detection is firstly presented in text detection field. Textboxes \cite{liao2017textboxes} presents a novel SSD-based text detection method, which adapts the size and aspect ratio of anchor and uses $1\times5$ convolutional filters for long text lines. Textboxes++ \cite{liao2018textboxes++} is based on textboxes but directly regresses the 8 vertices of the oriented bounding box. \cite{liu2017deep} designs rules for calculating the order of the vertices of the oriented bounding box, and proposes parallel IoU computation for timesaver. \cite{ma2018arbitrary} presents rotation region proposal networks (RRPN) that proposes oriented bounding boxes in RPN stage and uses Rotation Region-of-Interest (RRoI) pooling layer in R-CNN stage. The aspect ratio of text lines varies greatly, and limited anchors cannot cover the size or aspect ratio of all objects; thus, many methods are anchor-free. Both \cite{zhou2017east} and \cite{he2017deep} generate labels with shrunk segmentation maps, and regress the vertices or angles of the bounding box on positive pixels. \cite{lyu2018multi} generates a corner map and a position-sensitive segmentation map, calculates oriented bounding boxes based on the corner map, and calculates the score for each bounding box using the position-sensitive segmentation map. \cite{zhong2018anchor} presents anchor-free region proposal network (AF-RPN) based on Faster R-CNN with the same design as FPN \cite{lin2017feature}, and the label is calculated from the shrunk segmentation map instead of the anchors.

Horizontal bounding boxes cannot closely surround objects in aerial images, so the academic community begins to pay attention to oriented bounding box detection in aerial images. \cite{Xia_2018_CVPR} labels a large-scale dataset which contains 15 categories and 188282 instances, each labeled with an arbitrary quadrilateral (8 vertices). A novel detector which directly regresses 8 vertices based on Faster R-CNN is also presented. ICPR ODAI \cite{ICPR_ODAI} and CVPR DOTA \cite{CVPR_DOTA} competitions are organized based on this dataset. \cite{ding2018learning} presents a two-stage R-CNN method with RoI Transformer, which, in the first step, proposes horizontal bounding boxes. The first R-CNN outputs oriented bounding boxes, and the inputting of the second R-CNN are oriented bounding boxes. \cite{multi2019object} proposes a novel method named rotatable region-based residual
network (R3-Net) which can detect multi-oriented vehicles in aerial images and videos. The rotatable region proposal network (R-RPN) is adopted to generate rotatable region of interests (R-RoIs) which crops rotated rectangle areas from feature maps.

\section{Method}
\subsection{Overview}
Our method's pipeline is shown in Fig. \ref{framework}. Recently, anchor-free methods \cite{huang2015densebox} \cite{redmon2016you} \cite{wang2019region} \cite{zhong2018anchor} are widely used in object detection. In this study, we also use anchor-free RPN. In particular, the label of RPN is not calculated based on the overlap between the anchor and the ground truth; instead, the label is generated from the shrunk segmentation map of the oriented bounding box. Unlike traditional object detection tasks, the output bounding box of RPN is oriented, so a novel angle embedding method is adopted to better represent oriented bounding boxes. Segmentation maps with 8 channels (x, y, w, h, angle embedding) are generated in RPN stage. Then our model proposes oriented bounding boxs with Rotated RoIAlign in R-CNN stage, where Cascade R-CNN is used. In the first R-CNN, a novel IoU calculation method named length independent IoU (LIIoU) is adopted. To make IoU independent of the length of target box, we intercept part of the long side of the target box to obtain the maximum IoU between the proposed box and the intercepted target box. In this way, some long boxes will also have corresponding positive samples. Traditional IoU calculation method is used in the second R-CNN. The backbone of our network is based on FPN \cite{lin2017feature}, and we augment the network in the same way as PANet \cite{liu2018path} by adopting bottom-up Path Augmentation and feature fusion. Next, we will introduce each component in detail.

\subsection{Network Design}

\begin{figure}
  \begin{center}
       \includegraphics[width=1.0\linewidth]{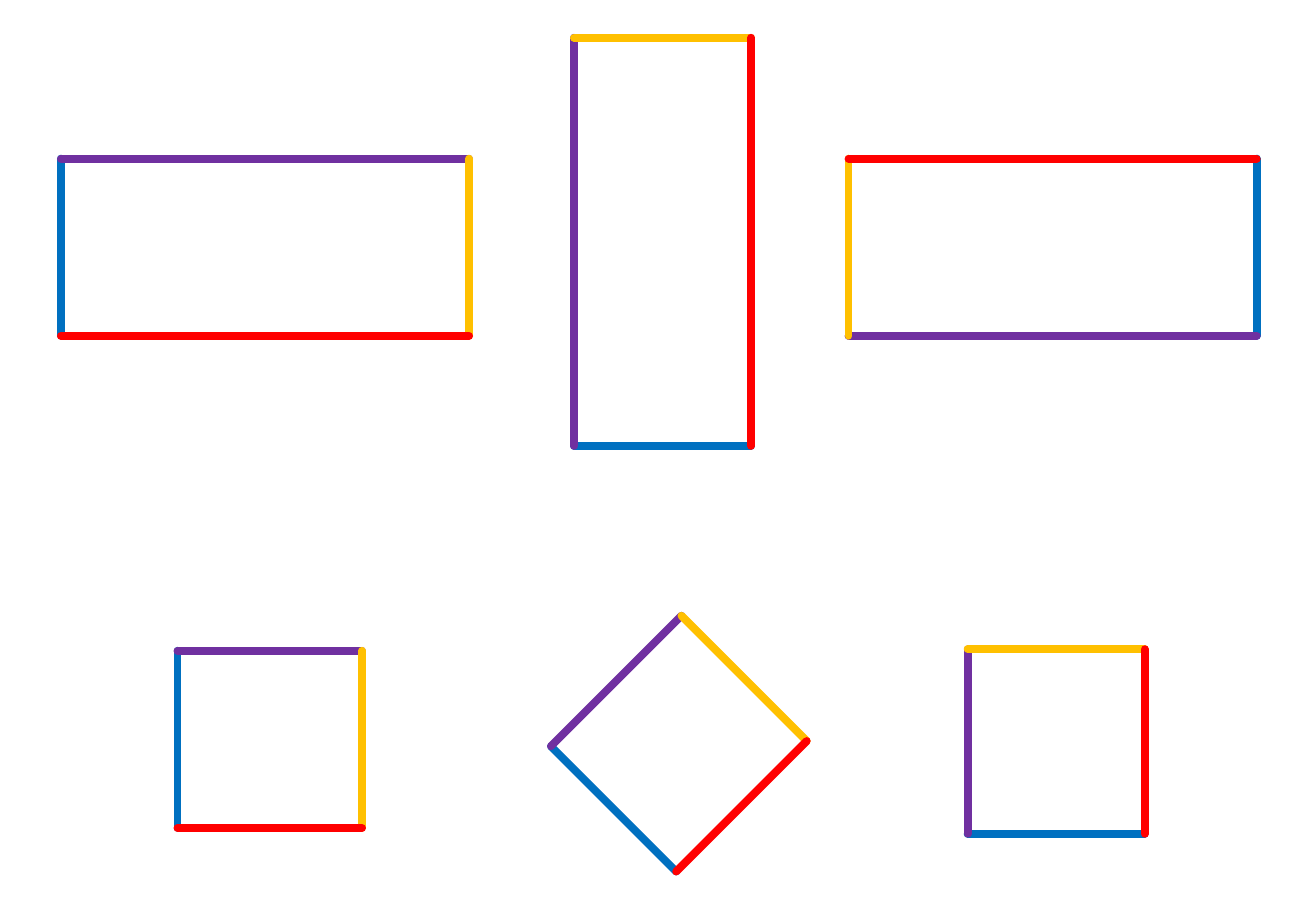}
    \end{center}
    \caption{The peroid of oriented bounding box. Top: rectangle whose peroid is $180^{\circ}$, down: squre whose peroid is $90^{\circ}$. (bounding box's sides are in different colors for better visualization)
  }
  \label{period}
\end{figure}

Inspired by recent object detection works \cite{lin2017feature} \cite{he2017mask} \cite{liu2018path}, we use Feature Pyramid Networks (FPN) as our backbone. FPN generates multiple feature maps with different sizes, and detect objects of different sizes on different layers. FPN is robust to scale variation expecially for small objects, which is suitable for this task. Besides scale variation, aspect ratio variation is another challenging problem. Most traditional object detection methods use anchors of different sizes and aspect ratios to calculate labels in RPN stage. Thus, we have to manually set the hyperparameter of the anchors which is too troublesome, and when the aspect ratio of objects varies greatly, limited anchors cannot cover all the objects. The performance of these detectors highly relies on anchor design. Recently, many methods \cite{huang2015densebox} \cite{redmon2016you} \cite{wang2019region} \cite{zhong2018anchor} \cite{zhou2017east} \cite{he2017deep} adopt the anchor-free strategy. In this study, we also adopt anchor-free method and generate the label of RPN from the shrunk segmentation map. Different layers extract different features, and the detector can achieve better performance by combining these features \cite{liu2018path}. In R-CNN stage, we fuse features of different layers after the first fully connected layer with max pooling.

\subsection{Anchor-free label generation}
\begin{figure}
  \begin{center}
       \includegraphics[width=0.8\linewidth]{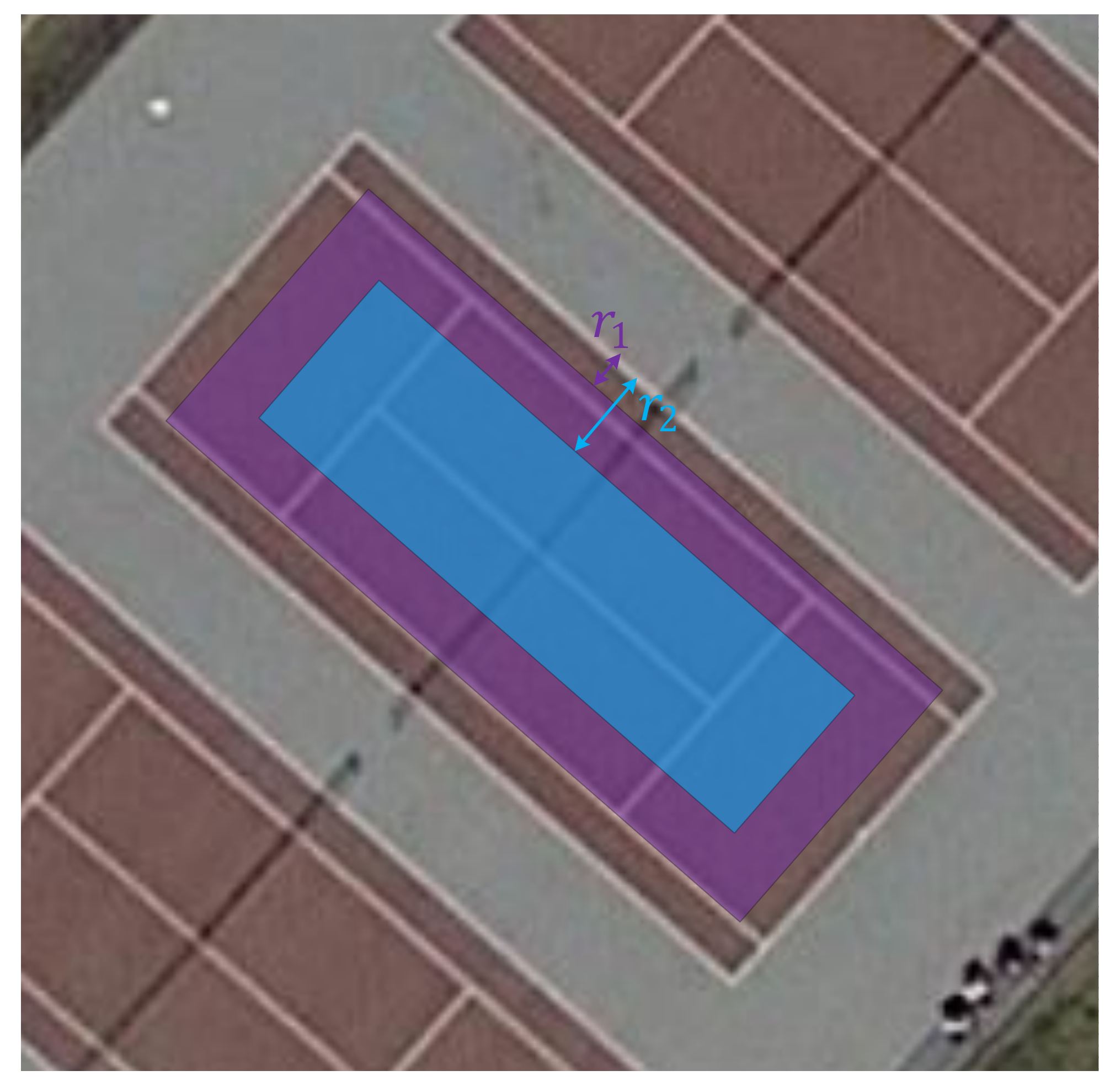}
    \end{center}
    \caption{The shrunk segmentation label of anchor-free method. Purple area is ignored area which is shrunk with $r_{1}$ ratio, blue area is positive area which is shrunk with $r_{2}$ ratio.
  }
  \label{anchor_free}
\end{figure}

Region proposals network (RPN) is adopted to propose candidate bounding boxes. Most of the previous methods are based on anchors in this stage. Considering the huge difference in aspect ratio, we use anchor-free RPN. The shrunk segmentation label is shown in Fig. \ref{anchor_free}. The shrinking method is the same as EAST \cite{zhou2017east}. In particular, $r_{1}$ is set to 0.1 and $r_{2}$ is set to 0.25. We shrink the oriented bounding box with $r_{2}$ ratio and set the pixels in the shrunk bounding box to positive samples (blue area). Next, we shrink the oriented bounding box with $r_{1}$ ratio, set the pixels in the shrunk bounding box but not set to positive samples (blue area) to ``do not care'' (purple area), and set the loss weight of these pixels to 0. FPN outputs multi-scale feature maps, and we detect objects of different scales on different layers. We assign a target object whose shorter side is $h$ to the level $p_{k}$, and $k$ is calculated as follows:

\begin{equation}
k= \lfloor k_{0} + \log_{2}(h/128) \rfloor
\end{equation}
where $k$ is the layer that objects should be assigned to; $k_{0}$ is the target layer when the height $h$ of the object is greater than 128 and less than 256, which we set to $4$. As the objects of different scales share the regression and classification parameters of RPN, the regression targets should be normalized. An oriented bounding box is labeled as:

\begin{equation}
(x_{\text{c}},y_{\text{c}},w,h,\theta)
\end{equation}
where ($x_{\text{c}}$, $y_{\text{c}}$) is the coordinates of the center point, $w$, $h$ are the lengths of the long side and the short side respectively, and $\theta$ is the angle of the long side. The pixel on $k$ layer is labeled as $x_{k},y_{k}$. First, we normalize the target bounding box with the stride of $k$ layer:

\begin{equation}
x^{\prime}_{\text{c}}=\frac{x_{\text{c}}}{s_{k}}, y^{\prime}_{\text{c}}=\frac{y_{\text{c}}}{s_{k}}, w^{\prime}=\frac{w}{s_{k}}, h^{\prime}=\frac{h}{s_{k}}, \theta^{\prime}=\theta
\end{equation}

where $s_{k}$ is the stride of $k$ layer calculated as:
\begin{equation}
s_{k}=2 \times 2^{k}
\end{equation}

The regression targets are calcualted as follows:

\begin{equation}
\begin{aligned}
t_{x_{\text{c}}}=\frac{x^{\prime}_{\text{c}}-x_{k}}{N},& \ \ \ t_{y_{\text{c}}}=\frac{y^{\prime}_{\text{c}}-y_{k}}{N}\\
t_{w}=\log \frac{w^{\prime}}{N},& \ \ \ \ t_{h}=\log \frac{h^{\prime}}{N}\\
\end{aligned}
\end{equation}

where N is a constant and is set to 6 as default.

\subsection{Adaptive Period Embedding}

A horizontal bounding box can be easily represented by 4 variables $(x, y, w, h)$. But we need an extra variable $\theta$ to represent an oriented bounding box. The primary challenge of oriented bounding box detection is to regress the angle of objects. The property of $\theta$ is different from other variables such as x, y, w, h, as $\theta$ is a periodic variable. As shown in Fig. \ref{period}, if the length and width of the rectangle are equal, the rectangle is a square, and the peroid of $\theta$ is $90^{\circ}$. Otherwise, the peroid of $\theta$ is $180^{\circ}$. In neural networks, the periodic property cannot be represented by one variable. Even though \cite{long2018textsnake} \cite{zhu2018textmountain} \cite{xu2019textfield} all use two-dimensional periodic vector $(\cos \theta, \sin \theta)$ for representing angle, they do not adapt vector's period. The proposed Adaptive Period Embedding (APE) uses two two-dimensional vectors to represent the angle. The the first vector has a period of $90^{\circ}$ and can be formulated as:
\begin{equation}
\mathbf{u}_{1}=(\cos 4\theta, \sin 4\theta)
\end{equation}
where $\theta$ is the angle of rectangle's long side. The period of the second vector is $180^{\circ}$ which represents the angle of rectangle's long side. It is calculated as follows:
\begin{equation}
\mathbf{v}=(\cos 2\theta, \sin 2\theta)
\end{equation}
\begin{equation}
\mathbf{u}_{2}=\mathbf{v} \times \min(\frac{(w-h)}{\lambda h},1)
\end{equation}

where $\lambda$ is set to 0.5, w is rectangle's long side, h is short side. Each component of $\mathbf{u}_{1}, \mathbf{u}_{2}$ is in $[- 1, 1]$, so we use sigmoid as activation, and then multiply them by 2 and subtract 1. Smooth L1 loss \cite{girshick2015fast} is used in all regression tasks of this study which can be formulated as:
\begin{align}\text{smooth}_{L_{1}}(z,z^{*})=
\begin{cases}
0.5(z-z^{*})^{2}& \text{if} \ |z-z^{*}| < 1\\
|z-z^{*}|-0.5& \text{otherwise}
\end{cases}
\label{smooth}
\end{align}

The final outputs of the neural network are $(x, y, w, h, \mathbf{u}_{1}, \mathbf{u}_{2})$. Next, we calculate the angle of the rectangle's long side based on $(\mathbf{u}_{1}, \mathbf{u}_{2})$. Firstly, $\theta_{90^{\circ}}$ whose peroid is $90^{\circ}$ can be calculated as:
\begin{equation}
\theta_{90^{\circ}}=\frac{atan2(\mathbf{u}_{1})}{4}
\end{equation}
where atan2 function calculates one unique arctangent value from a two-dimensional vector. The $\theta$ of rectangle's long side may be $\theta_{90^{\circ}}$ or $\theta_{90^{\circ}}+90^{\circ}$. The $\theta_{180^{\circ}}$ whose peroid is $180^{\circ}$ can be calculated as:
\begin{equation}
\theta_{180^{\circ}}=\frac{atan2(\mathbf{u}_{2})}{2}
\end{equation}

Then we calculate the distance between $\theta_{90^{\circ}}$ and $\theta_{180^{\circ}}$

\begin{equation}
\text{dis}=|(2 \theta_{90^{\circ}}-2 \theta_{180^{\circ}}+180^{\circ}) \mod 360^{\circ}-180^{\circ}|
\end{equation}
So the final $\theta$ is calculated as:
\begin{equation}
\theta=
\begin{cases}
\theta_{90^{\circ}} &  \text{dis}<90^{\circ} \\
\theta_{90^{\circ}}+90^{\circ} & \text{otherwise}
\end{cases}
\end{equation}

$\theta_{180^{\circ}}$ is the angle of the rectangle's long side. When the length and width of the rectangle are equal, the norm of $\mathbf{u}_{2}$ is nearly zero, so the angle calculated by $\mathbf{u}_{2}$ is not accurate. $\theta_{90^{\circ}}$ is accurate but it may be the angle of the long side or the short side. So we find the angle closer to $\theta_{180^{\circ}}$ from $\theta_{90^{\circ}}$ and $\theta_{90^{\circ}}+90^{\circ}$, which is the final result.

\subsection{length independent IoU}

\begin{figure*}
  \begin{center}
       \includegraphics[width=1.0\linewidth]{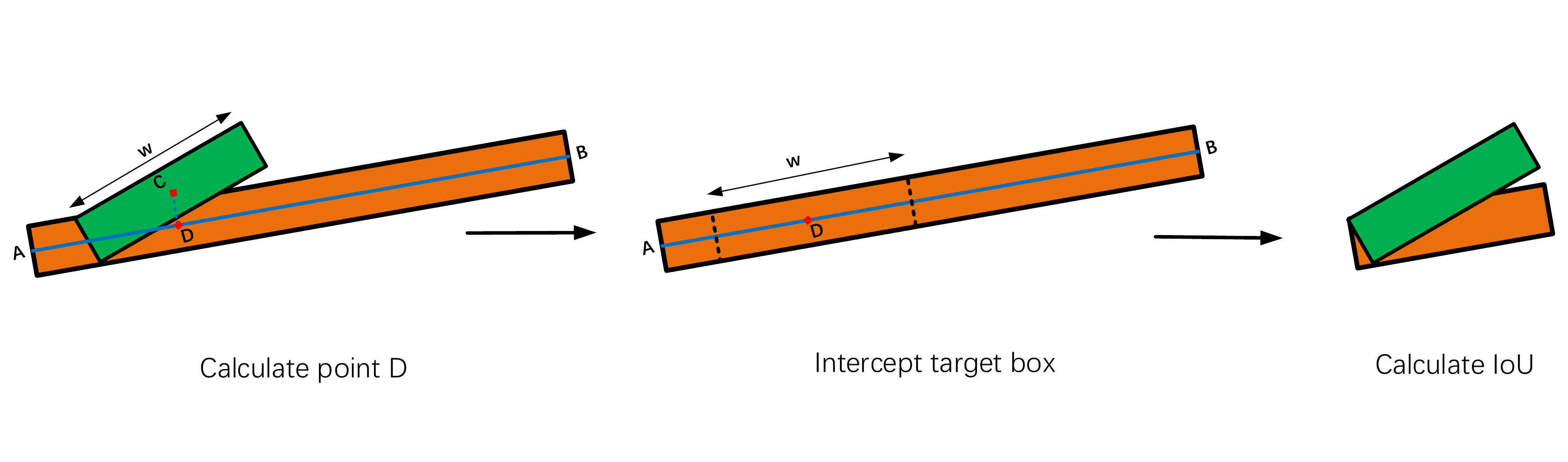}
    \end{center}
    \caption{The details of LIIoU calculation, green bounding box is proposed bounding box, orange
bounding box is target box. 
  }
  \label{LIIoU_cal}
\end{figure*}

IoU is the evaluation protocol of object detection, the more accurate the regression, the better the performance. But the receptive field of a neural network is limited and thus cannot cover some long objects. The detector proposes candidate bounding boxes in RPN, and then classifies and regresses these boxes again. The result of R-CNN highly relies on the output bounding boxes of RPN. In R-CNN stage, only the proposed bounding boxes whose IoU is higher than 0.5 is set to positive samples. Some target objects that are not well regressed in RPN cannot be detected in R-CNN. One idea is multiple regression \cite{cai2017cascade} in R-CNN stage, but if there are no positive proposed bounding boxes in the first R-CNN, the improvement is limited. Considering this, we propose a novel IoU calculation method named length independent IoU (LIIoU). We intercept part of the target box along its long side, and make the length of the intercepted box the same as the proposed box. The presented method is inspired by Seglink \cite{shi2017detecting}, but in our method, the aspect ratio of the proposed bounding box is arbitrary. As shown in Fig. \ref{framework}, the traditional IoU is only 0.3, but our proposed LIIoU is nearly 1. The details of LIIoU calculation is illustrated in Fig. \ref{LIIoU_cal}, where AB is the center line of the target box, and point C is the center of the proposed box. First, we find the perpendicular of AB through point C and label the intersection of the perpendicular and AB as point D. Next, we intercept a rectangle from the target bounding box as follows: if the length of the target box is smaller than the proposed box, we do not intercept; otherwise, the center of the intercepted rectangle is point D and the length is the same with proposed box (green box). Finally, we calculate IoU between the intercepted target box and the proposed box. The procedure is summarized in Algorithm~\ref{LIIOU_al}. In this way, more bounding boxes will regress targets in R-CNN which can improve the overall quality of the bounding boxes

\begin{algorithm}[h]
\caption{LIIoU calculation}

 {\textbf{Input}: $pbbox (x^{\text{p}},y^{\text{p}},w^{\text{p}},h^{\text{p}},\theta^{\text{p}})$, $gbbox (x^{\text{g}},y^{\text{g}},w^{\text{g}},h^{\text{g}},\theta^{\text{g}})$} \\
$pbbox$ - proposed bounding box\\
$gbbox$ - ground truth bounding box\\

\textbf{Output}: LIIoU\\
\begin{algorithmic}[1]

\IF{$w^{\text{p}}>=w^{\text{g}}$}
\STATE $x'^{\text{g}}=x^{\text{g}};y'^{\text{g}}=y^{\text{g}};w'^{\text{g}}=w^{\text{g}};h'^{\text{g}}=h^{\text{g}};\theta'^{\text{g}}=\theta^{\text{g}}$
\ELSE
\STATE $\mathbf{A}_{x}=x^{\text{g}}-\cos(\theta^{\text{g}})\times \frac{w^{\text{g}}}{2}$
\STATE $\mathbf{A}_{y}=y^{\text{g}}-\sin(\theta^{\text{g}})\times \frac{w^{\text{g}}}{2}$
\STATE $\mathbf{B}_{x}=x^{\text{g}}+\cos(\theta^{\text{g}})\times \frac{w^{\text{g}}}{2}$
\STATE $\mathbf{B}_{y}=y^{\text{g}}+\sin(\theta^{\text{g}})\times \frac{w^{\text{g}}}{2}$
\STATE $\mathbf{C}_{x}=x^{\text{p}};\mathbf{C}_{y}=y^{\text{p}}$
\STATE $z=\frac{(\mathbf{C}-\mathbf{A})\cdot(\mathbf{B}-\mathbf{A})}{||(\mathbf{B}-\mathbf{A})||}$
\STATE $w_{1}=z-\frac{w^{\text{p}}}{2};w_{2}=z+\frac{w^{\text{p}}}{2}$
\IF{$w_{1}<=0$}
\STATE $w_{1}=0;w_{2}=w^{\text{p}}$
\ELSIF{$w_{2}>=w^{\text{g}}$}
\STATE $w_{2}=w^{\text{g}};w_{1}=w^{\text{g}}-w^{\text{p}}$
\ENDIF
\STATE $x'^{\text{g}}={A}_{x}+\cos(\theta) \times \frac{w_{2}+w_{1}}{2};y'^{\text{g}}={A}_{y}+\sin(\theta) \times \frac{w_{2}+w_{1}}{2}$
\STATE $w'^{\text{g}}=w_{2}-w_{1};h'^{\text{g}}=h^{\text{g}};\theta'^{\text{g}}=\theta^{\text{g}}$
\ENDIF
\STATE calulate overlaps between $(x^{\text{p}},y^{\text{p}},w^{\text{p}},h^{\text{p}},\theta^{\text{p}})$ and $(x'^{\text{g}},y'^{\text{g}},w'^{\text{g}},h'^{\text{g}},\theta'^{\text{g}})$
\end{algorithmic}
\label{LIIOU_al}
\end{algorithm}

\subsection{Cascade R-CNN}

 As shown in Fig. \ref{framework}, two R-CNNs are used after RPN. In the first R-CNN, we only refine the center, height and width of the oriented bounding box without regressing the vertices of the target box. This is because the output of the first R-CNN is the input of the second R-CNN, and Rotated RoIAlign can only handle oriented rectangle but not quadrangle. In the second R-CNN, we regress the vertices of the target box.

Rotated RoIAlign is adopted, so the ground truth is calculated in a rotated coordinate system. If the center of a Rotated RoIAlign is $(x^{\text{p}}_{\text{c}},y^{\text{p}}_{\text{c}})$ and the angle is $\theta^{\text{p}}$, the affine transformation can be represented by an affine matrix:
\begin{equation}
 \begin{split}
\pmb{M}=&\left[
 \begin{matrix}
   1 & 0 & x^{\text{p}}_{\text{c}} \\
   0 & 1 & y^{\text{p}}_{\text{c}} \\
   0 & 0 & 1
  \end{matrix}
  \right]*
  \left[
 \begin{matrix}
   \cos \theta^{\text{p}} & \sin \theta^{\text{p}} & 0 \\
   -\sin \theta^{\text{p}} & \cos \theta^{\text{p}} & 0 \\
   0 & 0 & 1
  \end{matrix}
  \right]*
  \left[
 \begin{matrix}
   1  & 0 & -x^{\text{p}}_{\text{c}} \\
   0  & 1 & -y^{\text{p}}_{\text{c}} \\
   0 & 0 & 1
  \end{matrix}
  \right]\\
  =&
  \left[
 \begin{matrix}
   \cos \theta^{\text{p}}  & \sin \theta^{\text{p}} & (1-\cos \theta^{\text{p}})x^{\text{p}}_{\text{c}}-y^{\text{p}}_{\text{c}}*\sin \theta^{\text{p}} \\
   -\sin \theta^{\text{p}} & \cos \theta^{\text{p}} & (1-\cos \theta^{\text{p}})y^{\text{p}}_{\text{c}}+x^{\text{p}}_{\text{c}}*\sin \theta^{\text{p}}\\
   0 & 0 & 1
  \end{matrix}
  \right]
\end{split}
\label{eq3}
\end{equation}

\begin{equation}
\left(
 \begin{matrix}
   x \\
   y \\
   1
  \end{matrix}
  \right)=\pmb{M}
  \left(
  \begin{matrix}
   x' \\
   y' \\
   1
  \end{matrix}
  \right)
\label{eq4}
\end{equation}
We set the coordinate system to rotated coordinate system with Eq. \ref{eq3} and Eq. \ref{eq4}. The final ground truth in the rotated coordinate system is $(x,y)$, and $(x',y')$ is the coordinates in the original coordinate system. In the first R-CNN, the regression targets are $(t_{x_{\text{c}}}^{\text{rcnn}1},t_{y_{\text{c}}}^{\text{rcnn}1},t_{w}^{\text{rcnn}1},t_{h}^{\text{rcnn}1})$ which can be formulated as:

\begin{equation}
t_{x_{\text{c}}}^{\text{rcnn}1}=\frac{x_{\text{c}}-x^{\text{p}}_{c}}{w^{\text{p}}};t_{y_{\text{c}}}^{\text{rcnn}1}=\frac{y_{\text{c}}-y^{\text{p}}_{c}}{h^{\text{p}}}
\end{equation}

\begin{equation}
t_{w}^{\text{rcnn}1}=\log (\frac{w}{w^{\text{p}}});t_{h}^{\text{rcnn}1}=\log (\frac{h}{h^{\text{p}}})
\end{equation}

In the second R-CNN, the regression targets are $(t_{x_{i}}^{\text{rcnn}2},t_{y_{i}}^{\text{rcnn}2}), i=1,2,3,4$ which can be formulated as:

\begin{equation}
t_{x_{i}}^{\text{rcnn}2}=\frac{x_{i}-x^{\text{p}}_{c}}{w^{\text{p}}};t_{y_{i}}^{\text{rcnn}2}=\frac{y_{i}-y^{\text{p}}_{c}}{h^{\text{p}}}, i=1,2,3,4
\end{equation}

where $(x_{\text{c}},y_{\text{c}},w,h)$ are the center, width and height of the ground truth, $(x_{i},y_{i})$ is the vertex of the ground truth bounding box, and $(x^{\text{p}}_{c},y^{\text{p}}_{c},w^{\text{p}},h^{\text{p}})$ are the center, width and height of the proposed bounding box.

\subsection{Experiment}

\subsubsection{Datasets}

DOTA \cite{Xia_2018_CVPR} is a large dataset which contains 2806 aerial images from different sensors and platforms. The size of the image varies greatly, ranging from about $800 \times 800$ to $4000 \times 4000$ pixels, so it is necessary to crop the image and detect the objects in the cropped images. As the instances in arial images are oriented such as car, ship and bridge, each instance is labeled by an arbitrary (8 d.o.f.) quadrilateral. For the oriented task, the output bounding boxes are quadrilatera; to evaluate the performance of our detector on quadrilateral, we use the evaluation system provided along with this dataset. There are two versions of DOTA dataset, DOTA-v1.0 and DOTA-v1.5; DOTA-v1.5 fixes some errors and is provided for DOAI2019 competition \cite{CVPR_DOTA}. We use DOTA-v1.5 for this competition, but in the following experiments, we use DOTA-v1.0 for fair comparison.

\subsubsection{Implementation Details}

The backbone of our detector is ResNet-50 \cite{he2016deep} pre-trained on ImageNet \cite{deng2009imagenet}. The number of FPN channels is set to 256. In R-CNN stage, two fully connected (FC) layers are used, the channel of which is set to 1024. Feature fusion is applied after the first FC layer along with maxpooling. Batchnorm is not used in this study. Our network is trained with SGD, where the batchsize of images is $1$ and the initial learning rate is set to 0.00125, which is then divided by 10 at $\frac{2}{3}$ and $\frac{8}{9}$ of the entire training. Due to the limited memory, we crop images to $1024\times1024$ with the stride of $256$ for training and testing. The model is trained and tested at a single scale. Data augmentation is used for better performance; in particular, we randomly rotate images with angle among $0$, $\pi/2$, $\pi$, $3\pi/2$, and class balance resampling is adopted to solve class imbalance problem. In default, we train our model with training set and evaluate it on validation set and testing set. 

\subsubsection{Ablation Study}

In order to evaluate the effect of each component, we conduct abalation experiments on validation set of DOTA. The model is not modified except the component being tested. 
\begin{figure*}
  \begin{center}
       \includegraphics[width=1.0\linewidth]{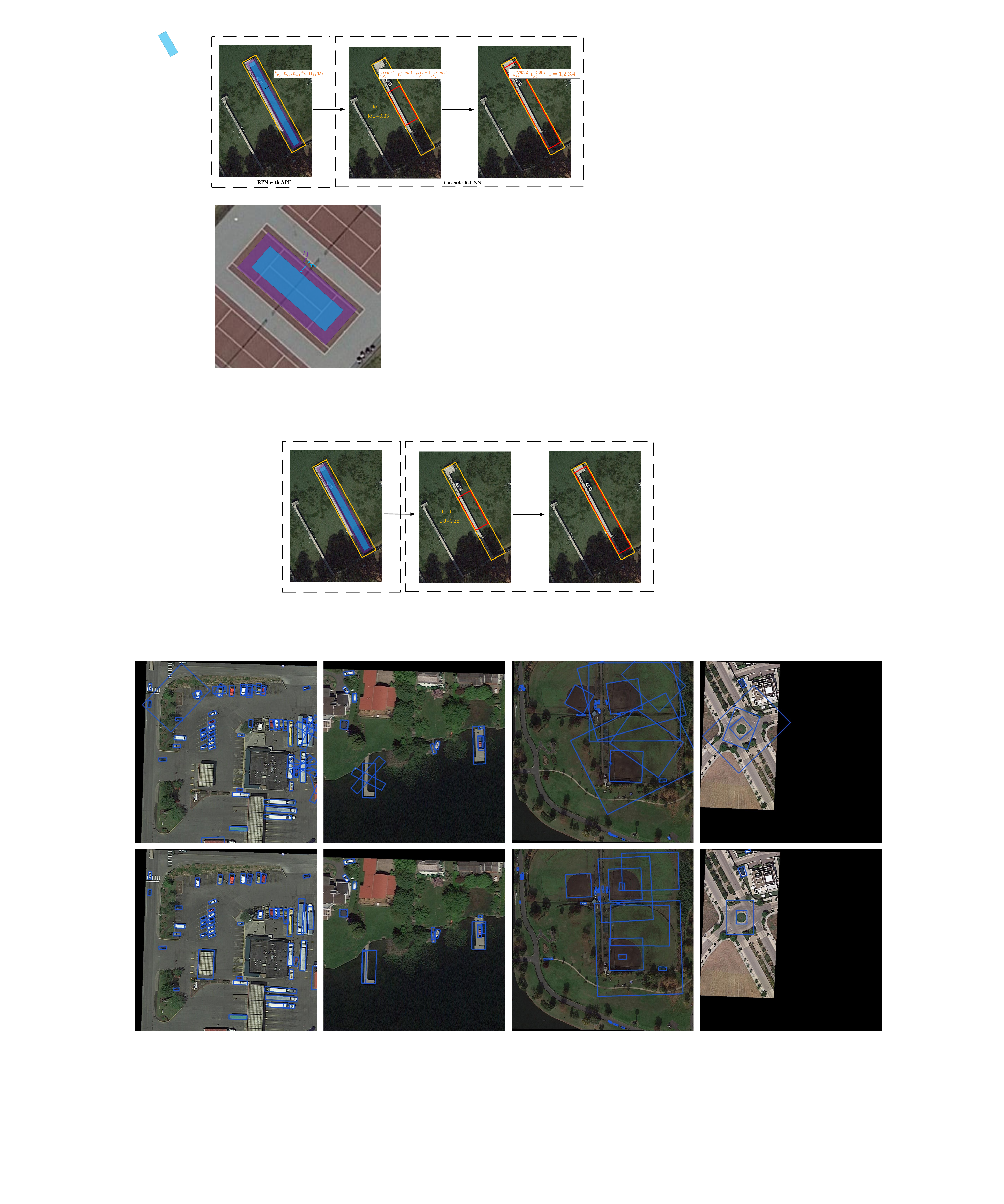}
    \end{center}
    \caption{The comparison of RPN with APE and without APE. top: without APE, down: with APE. 
    }
  \label{compare_APE}
\end{figure*}

\textbf{The effect of adaptive period embedding:} We need to propose oriented bounding boxes in RPN stage, but it is challenging to effectively represent a oriented bounding box. Most of previous methods \cite{zhou2017east} \cite{ding2018learning} \cite{ma2018arbitrary} which directly regress the angle do not notice the periodicity of the angle. When the angle is too diverse, the performance of the system will drop significantly. To evaluate whether the proposed APE can well handle the diversity of the angles, we conduct abalation experiments: one model directly regress the angle of the long side of the target box, while the other regresses adaptive period embedding (APE) vectors. We evaluate the quality of proposed oriented bounding box in RPN stage. Network only classifies objects into 2 classes (positive sample and negative sample) in RPN stage. As shown in Table \ref{APE_compare}, RPN achieves much better performance with APE. We show the comparison in Fig. \ref{compare_APE}, where we can see that RPN outputs more accurate angle with APE compared with directly regressing the angle. 

\begin{table}

  \begin{center}
      \caption{The experiment of APE on DOTA validation set in RPN stage. (in \%)}
    \begin{tabular}{|l|c|c|}
      \hline
      Methods &  w/o APE & w/ APE\\
      \hline
      AP  &70.16 & 72.20\\
       
      \hline

    \end{tabular}
  \label{APE_compare}
  \end{center}

\end{table}

\begin{table}
\centering
  \begin{center}
      \caption{The ablation experiments of LIIoU and IoU on DOTA validation set (in \%).}
    \begin{tabular}{|l|c|c|c|}
      \hline
      Methods &  Faster R-CNN & Cascade R-CNN & Cascade R-CNN+LIIoU \\
      \hline
      mAP  &71.40 &72.76 &\bf{73.88}\\
       
      \hline

    \end{tabular}
  \label{Ablation_liiou}
  \end{center}

\end{table}

\begin{table}

  \begin{center}
      \caption{Results on DOTA testing set (in \%). * indicates validation set is also used for training, otherwise only training set is used for training. }
    \begin{tabular}{|l|c|c|c|c|c|}
      \hline
      Method&Ours  & Ours * &FR-O \cite{Xia_2018_CVPR}   & RoI Transformer * \cite{ding2018learning} \\
      \hline
          Plane & 89.67 &  89.96 &79.09    &88.64 \\

          BD &76.77&  83.62  &69.12   &78.52\\

      Bridge &51.28 &  53.42  &17.17   &43.44 \\

      GTF & 71.65&  76.03  &63.49   & 75.92\\

      SV &73.11&  74.01  &34.2 &68.81 \\

      LV & 77.18&  77.16  &37.16   & 73.68 \\

      Ship &79.54 & 79.45  &36.2 &  83.59 \\

      TC &90.79 &  90.83 &89.19 &  90.74\\

      BC &79.01 & 87.15  &69.6 &77.27\\

      ST &84.54 &  84.51 &58.96   & 81.46 \\

      SBF & 66.51&  67.72 &49.4  &58.39 \\

      RA & 64.71& 60.33  &52.52   & 53.54 \\

      Harbor &73.97 & 74.61  & 57.79  &62.83 \\

      SP &67.73&  71.84 & 44.8  &58.93 \\

      HC &58.40&   65.55 &46.3  &47.67 \\

      \textbf{mAP} & 73.66 &  \textbf{75.75}  &52.93 &69.56 \\
      
      \hline

    \end{tabular}
  \label{compare_SOTA}
  \end{center}

\end{table}

\begin{table*}

  \begin{center}
      \caption{Task1 - Oriented Leaderboard on DOAI2019 (in \%).}
    \begin{tabular}{|l|c|c|c|c|c|c|}
      \hline
         Team Name  & USTC-NELSLIP  & \  pca\_lab \  & \ czh \ & \ AICyber \ & CSULQQ & \  peijin \   \\

      \hline
          Plane & 89.2&88.2 &89.0 &88.4&87.8&80.9\\

          BD &85.3& 86.4& 83.2& 85.4&83.6&83.6 \\

      Bridge &57.3 & 59.4&54.5 &56.7&56.7&55.1\\

      GTF & 80.9& 80.0& 73.8& 74.4&74.4&70.7\\

      SV &73.9& 68.1&72.6   & 63.9&63.2&59.9\\

      LV & 81.3& 75.6 & 80.3 & 72.7&71.0&76.4\\

      Ship &89.5 &87.2& 89.3 & 87.9&87.8&88.3 \\

      TC &90.8 &90.9 & 90.8 & 90.9&90.8&90.9\\

      BC &85.9 &85.3   &84.4 & 86.3&84.6& 79.2  \\

      ST &85.6 &84.1 & 85.0 & 85.0&84.0&78.3\\

      SBF & 69.5&73.8 &68.7   & 68.9&67.8&59.1\\

      RA & 76.7  &77.5 & 75.3 & 76.0&75.5&74.8\\

      Harbor &76.3 & 76.4& 74.2 & 74.1&67.4&74.1\\

      SP &76.0& 73.7 & 74.4 & 72.9 & 71.2 & 74.9 \\

      HC & 77.8 & 69.5 & 73.4 & 73.4 & 68.8 & 59.8 \\
      CC & 57.3 & 49.6 & 42.1 & 37.9 & 22.5 & 39.5 \\
      \textbf{mAP} & 78.3 &76.6 &75.7 & 74.7&72.3&71.6\\
      \hline

    \end{tabular}
  \label{DOAI2019}
  \end{center}

\end{table*}

\begin{figure}
  \begin{center}
       \includegraphics[width=1.0\linewidth]{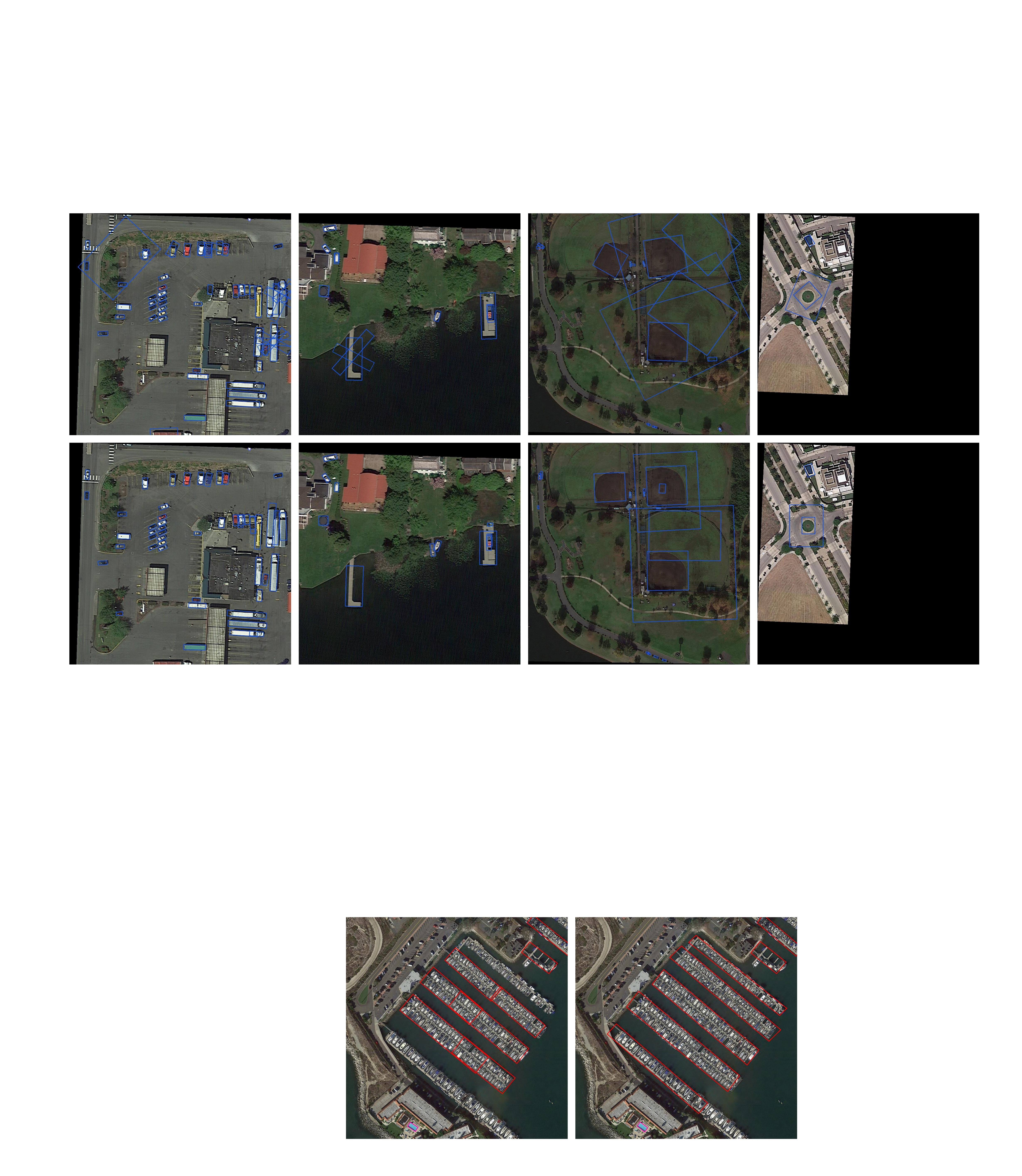}
    \end{center}
    \caption{The comparison of LIIoU and IoU. From Left to right: calculate overlaps with IoU in the first R-CNN, calculate overlaps with LIIoU in the first R-CNN (overlaps are both calculated with IoU in the last R-CNN).  
    }
  \label{compare_liiou}
\end{figure}

\textbf{LIIoU vs. IoU:} To evaluate the efficiency of LIIoU, we conduct control experiment. Faster R-CNN means there is only one R-CNN. When Cascade R-CNN is adopted, two R-CNNs are used. In the first model, we calculate the overlap between oriented bounding boxes with traditional IoU in both two R-CNNs. In the second model, the overlap between oriented bounding boxes is calculated with LIIoU in the first R-CNN and with traditional IoU in the last R-CNN, and the threshold is set to 0.5. Results are shown in Table \ref{Ablation_liiou}, where we can see that Cascade R-CNN gains much better performance with LIIoU. We show their comparison in Fig. \ref{compare_liiou}, where we can find that LIIoU can improve the quality of the proposed bounding boxes and the recall rate. Regardless of the aspect ratio and size, nearly every object has positive samples with LIIoU, so the detector can handle objects with large aspect ratios and lengths well.

\subsubsection{Comparing with other state-of-the-art methods}

\begin{figure*}
  \begin{center}
       \includegraphics[width=1.0\linewidth]{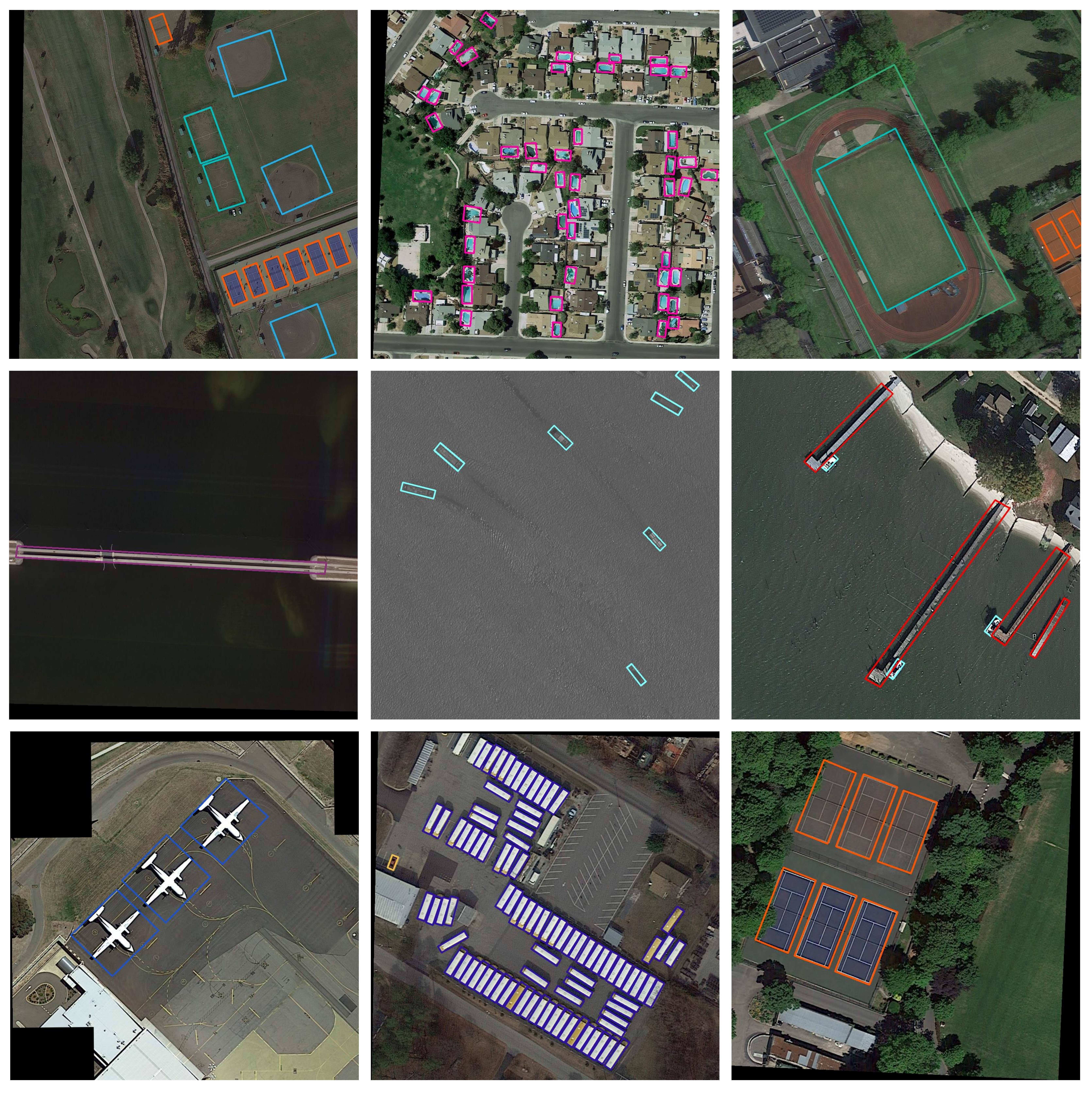}
    \end{center}
    \caption{Some results of our method on DOTA. The image's size is $1024 \times 1024$.
    }
  \label{results}
\end{figure*}

We compare our method with other state-of-the-art methods. The results are shown in Table \ref{compare_SOTA}. Our model is trained and tested with the single-scale setting. When our model is only trained with training set ex validation set, our method significantly outperforms other methods, if validation set is also used for training, it achieves better performance. The detection results are shown in Fig. \ref{results}. The angle, size, aspect ratio of objects in aerial images vary greatly, but our proposed method can well handle these challenging conditions.

\subsubsection{DOAI2019 competition}

DOAI2019 competition \cite{CVPR_DOTA} is held in workshop on Detecting Objects in Aerial Images in conjunction with IEEE CVPR 2019. The competition is more difficult and requires detecting all objects including samples labeled as difficult. Based on our proposed methods including APE and LIIoU, we adopt class balance resampling, image rotation, multi-scale training and testing and model assembling for better performance. Three models are used whose backbone is ResNeXt-101(32x4) \cite{xie2017aggregated}. Finally, we combine the training set with the validation set for training. The results of the competition are shown in Table \ref{DOAI2019}. Our method wins the first place on oriented task, with a gain of about 1.7\% over the most competing competitor.

\subsubsection{Conclusion and Future Work}

Detecting oriented objects in arial images is a challenging task. In this study, we make full use of the periodicity of the angle. A novel method named adaptive period embedding (APE) is proposed which can well regress oriented bounding boxes in arial images. The vector with adaptive period can learn the periodicity of the angle, which can not be implemented with one-dimensional vector. The proposed method can be applied to both one-stage methods such as RPN and two-stage methods, and we believe other detectors can also directly adopt APE module. Besides, we propose a novel length independent IoU (LIIoU). LIIoU set more proposed bounding boxes to positive samples expecially for long objects which can improve the quality of R-CNN regression. Our ablation study proves that each proposed module is effective. Our method achieves state-of-the-art on DOTA. Based on our method, we win the first place on oriented task of DOAI2019. In the future, we will explore more efficient and accurate detector for detecting oriented objects in arial images.

\ifCLASSOPTIONcaptionsoff
  \newpage
\fi



%
{
\bibliographystyle{IEEEtran}
\bibliography{refs}
}

\end{document}